\pdfoutput=1

\documentclass[11pt]{article}
\usepackage{adjustbox}
\usepackage[]{acl}
\usepackage{times}
\usepackage{forest}
\usepackage{latexsym}
\usepackage{amsmath}
\usepackage{booktabs}
\usepackage{multirow}
\usepackage{float}
\usepackage{listings}
\usepackage{adjustbox}
\usepackage{algorithm}
\usepackage{algorithmic}
\usepackage{amsmath}
\usepackage{amssymb}  
\usepackage{tabularx}   
\usepackage{longtable}
\usepackage{booktabs}   
\usepackage{multirow}   
\usepackage{array}  
\usepackage{siunitx}

\forestset{
  compact/.style={
    for tree={
      grow'=south,
      child anchor=north,
      parent anchor=south,
      align=center,
      l sep=3pt,
      s sep=3pt,
      edge={->, semithick},
      font=\ttfamily\scriptsize,
    },
  },
}
\usepackage[T1]{fontenc}

\usepackage[utf8]{inputenc}

\usepackage{microtype}

\usepackage{inconsolata}

\usepackage{graphicx}

\usepackage{xcolor}

\definecolor{update}{HTML}{E69F00}  
\definecolor{match}{gray}{0.25}     
\definecolor{insert}{HTML}{009E73}  

\title{SPENCE: A Syntactic Probe for Detecting Contamination in NL2SQL Benchmarks}

\author{
Mohammadtaher Safarzadeh\thanks{\texttt{mohammadtaher.safarzadeh@oracle.com}} \quad
Hitesh Laxmichand Patel \quad
Afshin Orojlooyjadid \\
\textbf{Graham Horwood} \quad \textbf{Dan Roth} \\  
Oracle AI
}
\begin{document}

\maketitle

\begin{abstract}
Large language models (LLMs) have achieved strong performance on natural language to SQL (NL2SQL) benchmarks, yet their reported accuracy may be inflated by contamination from benchmark queries or structurally similar patterns seen during training. We introduce \textbf{SPENCE} (\emph{Syntactic Probing and Evaluation of NL2SQL Contamination Effects}), a controlled syntactic probing framework for detecting and quantifying such contamination. SPENCE systematically generates syntactic variants of test queries for four widely used NL2SQL datasets—Spider, SParC, CoSQL, and the newer BIRD benchmark. We use SPENCE to evaluate multiple high-capacity LLMs under execution-based scoring. For each model, we measure changes in execution accuracy ($\Delta$~Accuracy) across increasing levels of syntactic divergence and quantify rank sensitivity using Kendall’s~$\tau$ with bootstrap confidence intervals. By aligning these robustness trends with benchmark release dates, we observe a clear temporal gradient: older benchmarks such as Spider exhibit the strongest negative 
$\tau$values, while the more recent BIRD dataset shows substantially weaker rank sensitivity. Together, these findings suggest that syntactic-probing evaluations can help identify contamination-like behavior and provide a more trustworthy view of NL2SQL generalization.
\end{abstract}

\section{Introduction}


\begin{figure}[t] 
\centering
\includegraphics[width=1\linewidth]{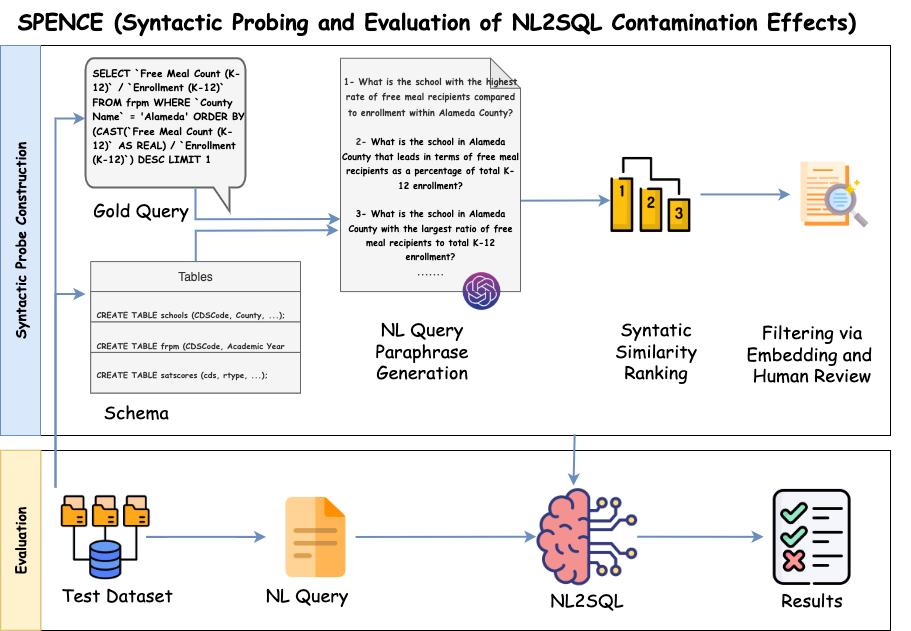}
\caption{Overview of the \textbf{SPENCE} pipeline for evaluating contamination in NL2SQL benchmarks. 
Given a test dataset, we extract the schema and gold SQL for each example and provide them, together with the natural-language (NL) query, to GPT-5 to generate ten syntactic paraphrases. 
The generated paraphrases are ranked by syntactic similarity using tree edit distance and filtered via embedding-based similarity and manual inspection to ensure semantic fidelity. 
Both the original and filtered paraphrased queries are then executed with the NL2SQL model, and execution accuracy is computed across paraphrase ranks. 
The resulting performance curves reveal how model robustness varies with syntactic divergence, enabling detection of contamination or memorization effects.}
\label{fig:flowchart}
\end{figure}

As NLP systems are increasingly deployed in real-world settings \citep{Singh2023}, reliable text-to-SQL systems are becoming an increasingly important practical application area \citep{singh-etal-2025-llms}. A growing body of work has examined data contamination and benchmark leakage in large language models.
\citet{dekoninck2024constatperformancebasedcontaminationdetection} highlight how benchmark scores can be artificially inflated and propose retrieval-based overlap checks and a ``testset slot guessing'' protocol to uncover hidden contamination in both open- and closed-source LLMs.
\citet{golchin2023dataquiz} and \citet{golchin2024timetravelllmstracing} introduce two complementary tools—the \emph{Data Contamination Quiz} and \emph{Time Travel in LLMs}—which convert benchmark items into multiple-choice or guided-completion tasks to reveal memorization signals at the instance and partition levels.
\citet{li2023opensource} compiles an open-source contamination report comparing overlap metrics, perplexity behavior, and n-gram matching across models and benchmarks.
\citet{oren2023proving} move beyond string-overlap and propose statistical tests based on canonical vs.\ shuffled benchmark orderings, providing provable guarantees of test-set contamination even in black-box settings.
\citet{shi2023detectpretraincode} and \citet{shi2023detecting} study the pretraining-data detection problem at the instance level, introducing open-source toolkits, the WikiMIA benchmark, and a simple Min-K\% Prob method that can flag member vs.\ non-member examples without knowing the training corpus.
On the same trend, \citet{mattern2023membership} and \citet{xu2024benchmarkleakage} provide broader benchmark-level studies, showing pervasive leakage across dozens of LLMs and recommending transparency practices such as ``Benchmark Transparency Cards.''
Together, these efforts demonstrate that contamination can occur in diverse ways and that performance drops on rephrased or reordered samples are a strong signal of non-generalizing, inflated benchmark scores—an insight our work builds upon with controlled paraphrase evaluation in the NL2SQL setting. 
For a comprehensive survey on data contamination see \citet{cheng2025surveydatacontaminationlarge, ravaut2025comprehensivesurveycontaminationdetection}.

We investigate whether LLMs exhibit contamination-like behavior on four widely used NL2SQL benchmarks—Spider \citep{Spider_paper}, SParC \citep{yu-etal-2019-sparc}, CoSQL \citep{yu-etal-2019-cosql}, and BIRD \citep{Bird_paper}—by probing the effect of syntactic paraphrasing of the natural-language queries on model performance. These benchmarks were released at different times: Spider in September 2018, SParC in June 2019, CoSQL in October 2019, and BIRD in May 2023. Given its earlier release and widespread adoption, we hypothesize that Spider is the most vulnerable to contamination effects, whereas the more recent BIRD benchmark is least likely to be affected. Throughout this paper we use ``contamination'' in an operational sense: behavior consistent with memorization or near-memorization of benchmark instances due to prior exposure---characterized by unusually high stability at small syntactic distances coupled with sharp degradation as structural divergence increases---rather than general compositionality failure, which would affect all queries more uniformly. Our experimental protocol applies controlled paraphrase transformations to each benchmark’s development set and quantifies the corresponding degradation in execution accuracy, providing an empirical measure of contamination risk across datasets.

\begin{figure*}
\centering
\includegraphics[width=\textwidth]{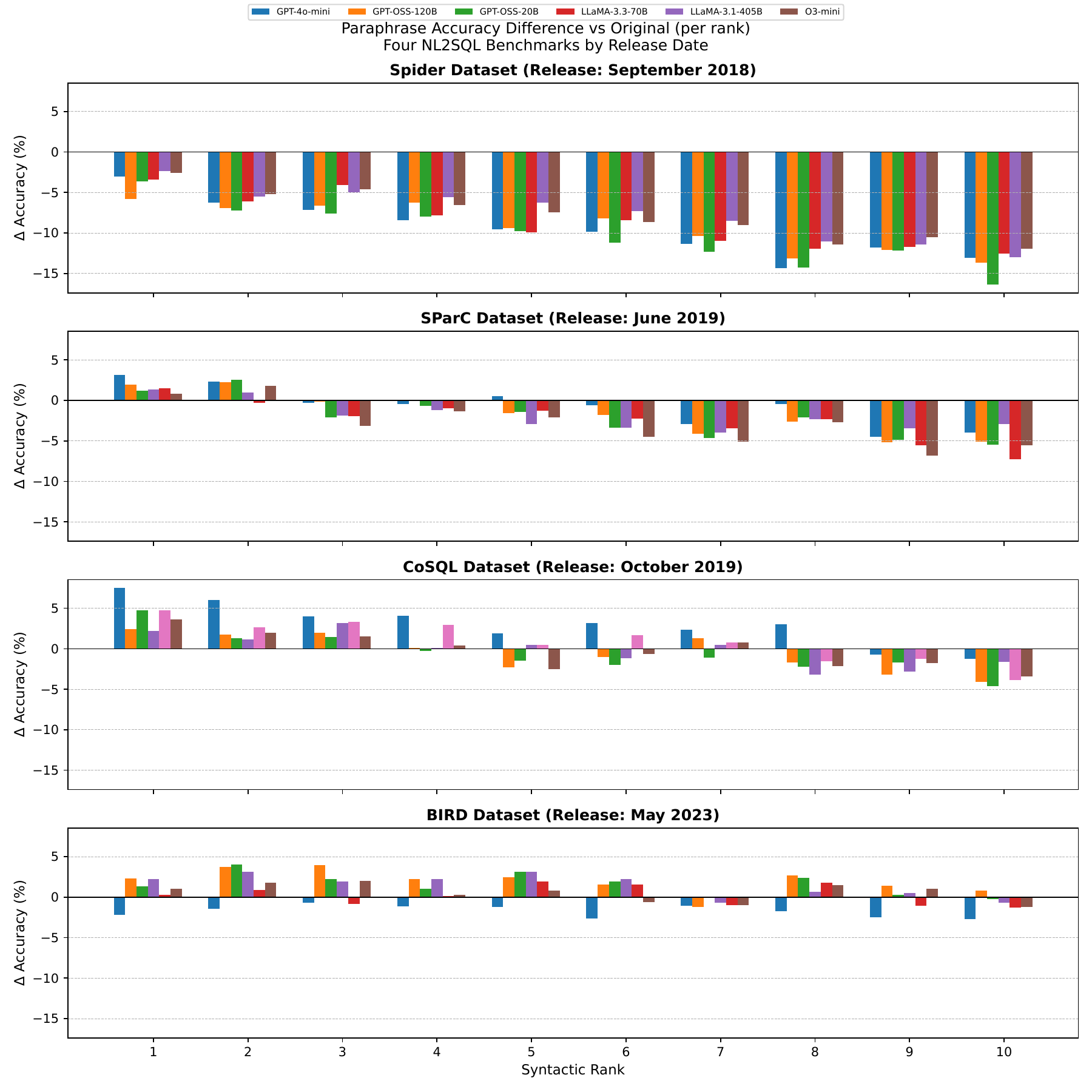}
\caption{Difference in execution accuracy (Paraphrase\,–\,Original queries) for each model across paraphrase ranks on four NL2SQL benchmarks — Spider (Sep 2018), SParC (Jun 2019), CoSQL (Oct 2019), and BIRD (May 2023) — shown in release order.
  Bars above the zero line indicate paraphrased queries performed better than the original queries on the same subset; bars below zero indicate performance degradation.
  All panels share the same $x$-axis (paraphrase rank) and $y$-axis scale (accuracy difference).}
\label{fig:accuracy-by-rank}
\end{figure*}

Our paraphrase–based probing shares conceptual similarities with the recent \emph{ConStat} framework \citet{dekoninck2024constatperformancebasedcontaminationdetection}, which redefines contamination in large language models as \emph{performance inflation that fails to generalize} rather than simple string overlap. 
ConStat introduces three contamination regimes depending on the choice of reference benchmark: \emph{syntax-specific} (paraphrased versions of the same examples), \emph{sample-specific} (new examples from the same underlying distribution), and \emph{benchmark-specific} (different but related benchmarks). 
\citet{dekoninck2024constatperformancebasedcontaminationdetection} report that sample-specific contamination is often more pronounced and easier to detect than syntax-specific contamination, indicating that models may memorize dataset artifacts beyond the surface form of questions. 
Our study focuses on the syntax axis—paraphrasing original NL queries to create increasingly distant rephrasings—but our methodology could naturally be extended with ConStat-style sample-specific probes (e.g., generating new but semantically equivalent NL–SQL pairs) to capture deeper forms of contamination. 
Thus our results complement ConStat’s findings: whereas ConStat emphasizes detecting non-generalizing performance across benchmarks, our approach offers a fine-grained view of generalization across controlled syntactic transformations.

Our work also complements recent investigations into contamination in text-to-SQL models. 
\citet{ranaldi2024investigating} study GPT-3.5’s performance on the Spider benchmark versus a newly introduced, presumably unseen dataset called \emph{Termite}. 
They propose several probes—most notably an \emph{adversarial table disconnection} (ATD) manipulation and a \emph{DC-accuracy} test—to measure how much a model has memorized schema and content. 
Their results show large performance drops on Termite and under ATD, suggesting that high scores on Spider may in part reflect \emph{sample-level contamination} rather than true generalization. 
Whereas their analysis operates at the dataset level by contrasting seen versus unseen databases, our approach focuses on the \emph{syntactic axis}: creating increasingly distant paraphrases of original NL queries while holding schemas fixed. 
Together, these perspectives give a more complete view of contamination, spanning both \emph{sample-level} and \emph{syntax-level} generalization.

\section{Methods}
\label{sec:methods}

We introduce \textbf{SPENCE} (\emph{Syntactic Probing and Evaluation of NL2SQL Contamination Effects}), 
a controlled framework for detecting and quantifying contamination in natural language to SQL (NL2SQL) benchmarks. 
SPENCE applies systematic syntactic perturbations to benchmark queries to evaluate whether large language models (LLMs) rely on surface-form memorization or exhibit genuine compositional generalization. 
Figure~\ref{fig:flowchart} illustrates the end-to-end evaluation pipeline, which proceeds from the generation of syntactic variants to execution-based assessment and rank-sensitivity analysis across models and datasets. 
We describe each component of the framework in detail below.

\subsection{Data Selection and Preprocessing}
Our objective is to test whether NL2SQL models exhibit contamination‐like behavior on public benchmarks under a realistic ``black box'' setting where we only observe outputs, not parameters or training data \citep{brown2020language,touvron2023llama2}. 
We conduct all experiments on the development splits of four widely used NL2SQL benchmarks: 
\textbf{Spider}, \textbf{BIRD}, \textbf{CoSQL}, and \textbf{SParC}. 
For each benchmark, we extract the \emph{gold SQL queries} and associated \emph{database schemas}, providing a clean ground-truth mapping between natural-language (NL) questions and their executable SQL. For the conversational benchmarks SParC and CoSQL, we paraphrase only the final user question in each dialogue, while keeping all preceding turns unchanged so that the dialogue state and grounding context are preserved. Execution-accuracy changes across paraphrase ranks can therefore be attributed more directly to syntactic perturbations of the target query, rather than to changes in conversational context.

\subsection{Paraphrase Generation and Filtering}

For our NL2SQL experiments we began with the test set from the Spider dataset and created \textbf{10 paraphrased versions} of each NL query using large language models (LLMs). 
To ensure the generated queries were valid and semantically meaningful, we included the \emph{gold SQL query} and its associated \emph{database schema} in the prompt. 
This step minimizes invalid or ill-formed paraphrases.

Each generated paraphrase was parsed into a syntactic dependency tree using the \texttt{spaCy} package (v3.7)\footnote{\url{https://spacy.io}}. 
We then computed the \emph{tree edit distance (TED)} between each paraphrase and the original NL query, where the allowed operations are insertions, deletions, and substitutions of \emph{dependency-tree nodes} (not just individual tokens). 
TED thus measures the minimal number of such operations needed to transform one parse tree into another and serves as our syntactic similarity metric. An example of this is provided in Appendix \ref{app:ted}.

Paraphrases were \emph{ranked by TED score} from 1 (closest to the original query) to 10 (most distant). 
Because absolute TED ranges can differ across base queries, the rank index indicates relative distance within a single query but not across queries. 
To further filter noise, we computed an \emph{embedding-based cosine similarity} \citep[E5-base-v2; ][]{wang2024textembeddingsweaklysupervisedcontrastive} and retained only paraphrases above a manually selected threshold (chosen via pilot inspection for clear separation between meaningful and irrelevant paraphrases). 
This yields nested “grades” of increasingly distant paraphrases for probing model generalization (see Appendix \ref{app:para_example} Table~\ref{tab:ted-71-ruled} for statistics on retained paraphrases).
To quantitatively verify that this filtering yields semantically faithful paraphrases across rank tiers, we additionally conduct a small human validation study on 100 Spider examples spanning ranks 1, 5, and 10. Over 85\% of paraphrases are judged semantically equivalent to their originals, with no systematic variation across rank
(Appendix~\ref{app:human-eval}).

Low TED scores typically correspond to superficial lexical changes (synonym swaps, minor word reordering), whereas high TED scores capture more substantial structural changes (clause reordering, subquery restructuring).

\begin{algorithm}[t]
\caption{Paraphrase Generation and Filtering}
\label{alg:paraphrase}
\begin{algorithmic}[1]
\FOR{each NL query $q$ in Spider test set}
  \STATE Generate 10 paraphrases $\{p_i\}_{i=1}^{10}$ using LLM with $(q_{\text{goldSQL}}, \text{schema})$ as context
  \FOR{each paraphrase $p_i$}
    \STATE Parse $p_i$ and $q$ into dependency trees $T_{p_i}, T_{q}$ using \texttt{spaCy}
    \STATE Compute TED$_i = \text{TreeEditDistance}(T_{p_i}, T_{q})$
    \STATE Compute cosine similarity $c_i = \text{CosSim}(\text{Embed}(p_i), \text{Embed}(q))$
  \ENDFOR
  \STATE Rank $\{p_i\}$ by TED$_i$ (ascending)
  \STATE Retain paraphrases with $c_i \ge \tau$ (manually chosen threshold)
\ENDFOR
\end{algorithmic}
\end{algorithm}

\subsection{Evaluation on Original and Paraphrased Queries}

Once the final paraphrase sets were obtained, we evaluated model performance on the original and paraphrased queries in a paired manner. 
For each base question, we took its $k$-th paraphrase (i.e., the paraphrase with rank $k$ according to TED) and compared the model’s output on that paraphrase to its output on the \emph{same underlying original question}. 
This ensures that at “rank $k$” we are always comparing paraphrases of the same original questions rather than shifting to a different question subset. 
Reporting results in this paired fashion isolates the effect of paraphrasing from any changes in the underlying question set.

We treat contamination detection as a paired hypothesis test:
\begin{itemize}
  \item $H_{0}$: the model’s accuracy is invariant to paraphrasing (no contamination / robust generalization);
  \item $H_{1}$: the model’s accuracy degrades with syntactic distance from original NL query because of memorization or leakage.
\end{itemize}

We evaluated the results using both reasoning and non-reasoning models and compared performance metrics across ranks. 
Interestingly, as syntactic distance increased (i.e., moving toward paraphrase rank 10), we observed a considerable drop in accuracy. 
This pattern suggests that models may be overly dependent on syntactic similarity in the NL query and can struggle with paraphrases more distant from the original—a possible sign of contamination or memorization effects.

\subsection{LLMs}
In order to get the paraphrased queries, we used GPT-5.0 \citep{openai2025gpt5}. We verify in Section~\ref{sec:critical} that our findings do not
depend on this specific choice of paraphrase generator. Then, for SQL query generation, we choose six LLMs, ranging from small to big open-source LLMs: GPT-OSS-20B, GPT-OSS-120B \citep{openai2025gptoss120bgptoss20bmodel}, LLaMA3.3-70B, LLaMA3.1 405B \citep{dubey2024llama}, and two commercial models: GPT-4o-mini \citep{openai2024gpt}, and GPT-O3-mini \citep{openai2025o3}. 

\subsection{Summary of Experimental Design}
Across all four benchmarks, our pipeline (i) generates controlled paraphrase sets at varying distances, (ii) filters them using cosine similarity thresholds, (iii) evaluates original vs.\ paraphrased accuracy for each subset, and (iv) compares models under the $H_{0}$/$H_{1}$ framing above. 
This design allows us to quantify not only how paraphrasing affects execution accuracy but also how potential data leakage influences a model’s robustness to linguistic variation.

In all our experiments, the results are provided by setting the sampling temperature to $T=1$. In Appendix \ref{app:temperature} we show that our results are immune to this by showing the results for $T=0$.

Finally, we note that the SPENCE pipeline is inexpensive to run. For a given benchmark, dependency parsing scales linearly in query length,
and tree edit distance between two dependency trees is computed in low-degree polynomial time using the standard Zhang--Shasha algorithm~\citep{zhang1989simple}; both are applied once per paraphrase and contribute negligibly to the total runtime. The dominant costs in the pipeline are the LLM calls for paraphrase generation and for downstream NL2SQL evaluation, which together account for essentially all of the wall-clock time. Parsing, ranking, and filtering are performed once per dataset and can be cached, so repeating the evaluation with additional downstream models does not incur additional
parsing or TED overhead.

\section{Results}

\subsection{Spider}
Top panel of Figure~\ref{fig:accuracy-by-rank} shows the results of our syntactic paraphrasing probe on the Spider benchmark. Performance on Spider drops sharply even for the earliest paraphrase ranks: all models incur noticeable accuracy losses from rank~1 onward, and the decline accelerates steadily as paraphrases become more syntactically distant. By ranks~8--10, the difference in execution accuracy exceeds 10--15 percentage points for every model tested. This pattern supports our hypothesis that Spider---as the earliest and most widely used NL2SQL benchmark (released in September 2018)---is the most vulnerable to contamination or overfitting to near-verbatim queries.

\subsection{SparC}
Second panel in Figure~\ref{fig:accuracy-by-rank} shows the results of our syntactic paraphrasing probe on the SParC-SQL benchmark. Across all six LLMs evaluated on the SParC-SQL benchmark, accuracy remains roughly stable or slightly improves for the first one to three paraphrase variants, which are closest in syntax to the original queries. Beginning at rank~4, however, the differences in accuracy turn negative and decline steadily as paraphrases become more syntactically distant. By ranks~8--10, all models exhibit substantial performance drops, in some cases exceeding 10--15 percentage points. Larger models such as LLaMA-3.1-405B degrade less sharply than smaller ones like o3-mini, but no model is immune. These results indicate that syntactic paraphrasing, especially at high divergence, exposes a hidden fragility in NL2SQL performance and provides an empirical signal of potential contamination or overfitting to near-verbatim queries.

\subsection{CoSQL}
Third panel in Figure~\ref{fig:accuracy-by-rank} shows the results of our syntactic paraphrasing probe on the CoSQL benchmark. Model performance is initially robust, with most systems exhibiting small positive gains for the first few paraphrase ranks, similar to what we observe on SParC. However, starting around rank~6, the differences in accuracy begin to turn negative and the drop accelerates toward higher paraphrase distances. By ranks~9--10, all models show clear degradation, though the magnitude of the drop is less severe than on Spider and somewhat more pronounced than on BIRD. This intermediate pattern is consistent with CoSQL’s release date (October 2019) and its position between the earliest and most widely used benchmark (Spider) and the much newer BIRD dataset: contamination effects appear weaker than on Spider but stronger than on BIRD, underscoring the value of CoSQL as a mid-point stress test for paraphrase robustness.

\subsection{BIRD}
Bottom panel of Figure~\ref{fig:accuracy-by-rank} shows the results of our syntactic paraphrasing probe on the BIRD benchmark. In contrast to Spider and SParC, model performance on BIRD remains remarkably stable under paraphrasing. For the first several paraphrase ranks, most models even show slight gains relative to the original queries, and the differences hover near zero across ranks~1--10. Only small, sporadic drops are observed for some models at higher ranks, and these never approach the steep degradations seen on the older benchmarks. This pattern is consistent with our expectation that BIRD---as the most recent NL2SQL benchmark (released in May 2023)---is least vulnerable to contamination or overfitting to near-verbatim queries, thereby providing a stronger measure of generalization under syntactic variation.

\begin{figure*}[t]
  \centering
  \begin{minipage}[t]{0.48\textwidth}
    \centering
    \includegraphics[width=\linewidth]{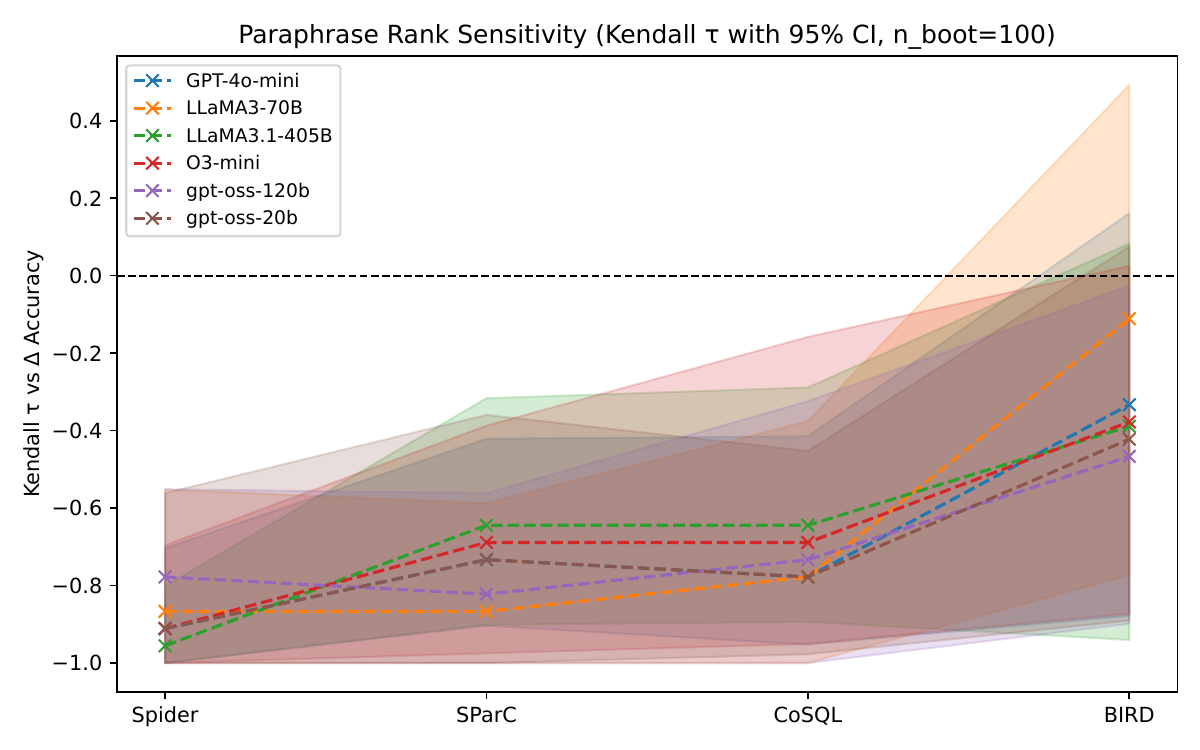}
    \textbf{(a) All paraphrase ranks.}
  \end{minipage}\hfill
  \begin{minipage}[t]{0.48\textwidth}
    \centering
    \includegraphics[width=\linewidth]{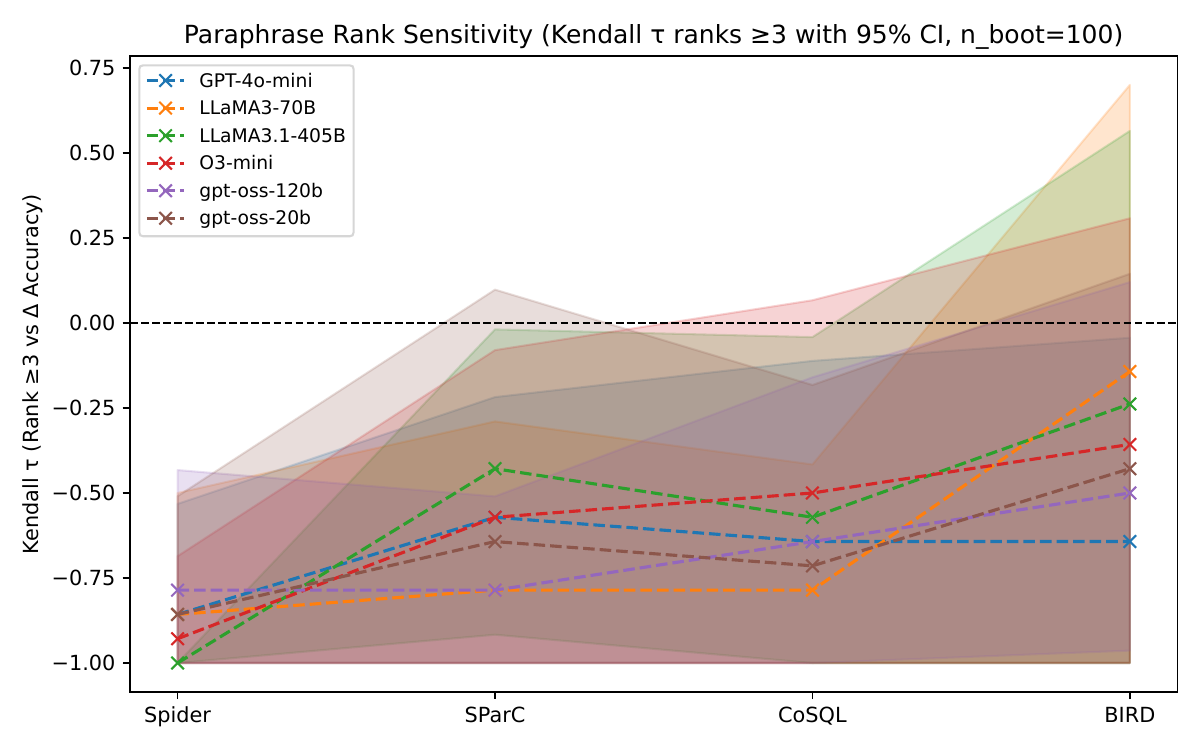}
    \textbf{(b) Paraphrase ranks $\geq 3$.}
  \end{minipage}

  \caption{Kendall’s~$\tau$ between paraphrase rank and $\Delta$ Accuracy (paraphrase accuracy minus original accuracy) across the four NL2SQL benchmarks (Spider, SParC, CoSQL, BIRD). 
  Each dashed line corresponds to a model, and the shaded region shows the 95\% bootstrap confidence interval based on 100 resamples. 
  (a)~All paraphrase ranks. (b)~Only ranks~$\geq 3$. Strongly negative $\tau$ values on Spider, SParC and CoSQL indicate sensitivity to paraphrase rank (accuracy drops for more distant paraphrases), whereas the near-zero values and wide confidence intervals on BIRD suggest little or no such sensitivity on the newer benchmark. See section \ref{sec:tau_trend} for details of the method implemented.} 
  \label{fig:kendall-two-panels}
\end{figure*}

\section{Results: Sensitivity of Models to Paraphrase Rank}
\label{sec:tau_trend}

To quantify how sensitive each model is to the ordering of paraphrased queries, we compute Kendall’s~$\tau$ correlation between paraphrase rank and the change in execution accuracy relative to the original queries ($\Delta$~Accuracy). 
Kendall’s~$\tau$ measures the ordinal association between two variables by comparing the number of concordant and discordant pairs in their rankings; formally,
\[
\tau = \frac{n_c - n_d}{\tfrac{1}{2}n(n-1)},
\]
where $n_c$ and $n_d$ denote the numbers of concordant and discordant pairs, respectively, and $n$ is the total number of observations. 
The coefficient ranges from $-1$ (perfect inverse correlation) to $+1$ (perfect agreement), with $0$ indicating no monotonic relationship. 

To estimate uncertainty, we apply non-parametric bootstrapping: for each model–dataset pair, we randomly resample the $n$ paraphrase–accuracy pairs with replacement $B = 100$ times, recompute Kendall’s~$\tau^{*(b)}$ for each resample, and derive the empirical 95\% confidence interval from the 2.5th and 97.5th percentiles of the bootstrap distribution $\{\tau^{*(b)}\}_{b=1}^{B}$. 
This approach provides a robust, distribution-free estimate of variability in the rank correlation. 
Positive $\tau$ values indicate that accuracy tends to increase with paraphrase rank, whereas negative values indicate a decline in accuracy as rank increases.

\paragraph{Trends across benchmarks.}
Figure~\ref{fig:kendall-two-panels} summarizes these correlations. Across the three older benchmarks—Spider, SParC, and CoSQL—all models show strongly negative Kendall’s~$\tau$ values (typically between $-0.6$ and $-0.95$), indicating a consistent degradation in accuracy as paraphrases become more syntactically distant. The bootstrap confidence intervals for these datasets do not overlap zero, confirming that this trend is statistically robust. In contrast, on the newer BIRD benchmark, Kendall’s~$\tau$ values cluster near zero (roughly $-0.46$ to $-0.11$, with most confidence intervals spanning zero), suggesting that model performance is largely insensitive to paraphrase rank.

This divergence between older and newer datasets supports the hypothesis that models may have memorized or been exposed to earlier benchmarks, yielding higher accuracy on paraphrases closest to the originals, whereas the newer BIRD dataset shows little evidence of such contamination. In other words, sensitivity to paraphrase rank appears to “wash out” for more recent benchmarks.

\paragraph{Comparison of all vs.\ higher-ranked paraphrases.}
To assess whether performance degradation is driven primarily by distant paraphrases, 
we recompute Kendall’s~$\tau$ using (i) all paraphrase ranks and (ii) only ranks~$\geq 3$. 
Table~\ref{tab:kendall-summary} reports the mean correlation across models for each dataset, 
while Table~\ref{tab:kendall-combined} in Appendix~\ref{app:kendall} provides full per-model results with 95\% bootstrap confidence intervals.

Across Spider, SParC, and CoSQL, the correlations remain strongly negative 
($\tau \approx -0.7$ to $-0.9$), confirming that execution accuracy consistently declines 
as syntactic divergence increases—even when excluding near-identical paraphrases. 
For these three datasets, the negative and statistically significant correlations persist 
when restricted to ranks~$\geq 3$ ($\tau$ between $-0.57$ and $-1.00$), indicating that 
the degradation is not confined to minor surface variations but extends to more substantive syntactic rewrites. 
In contrast, BIRD exhibits weak, statistically indistinguishable correlations under both settings ($\tau \approx -0.35$), 
suggesting limited rank sensitivity and reinforcing that contamination effects are more pronounced in older benchmarks 
but largely absent in newer data.

\begin{table}[t]
\centering
\caption{
Average Kendall’s~$\tau$ between paraphrase rank and $\Delta$~Accuracy across models for each dataset. 
Results are shown for all paraphrase ranks and for higher-ranked paraphrases ($\geq$3), 
isolating the effect of more distant paraphrases. 
See Appendix~\ref{app:kendall} for per-model results with confidence intervals.
}
\label{tab:kendall-summary}
\begin{tabular}{l @{\hspace{0.3cm}} c @{\hspace{0.3cm}} c}
\toprule
Dataset & All ranks & Ranks~$\geq$3 \\
\midrule
Spider & \textbf{-0.89} & \textbf{-0.88} \\
SParC  & \textbf{-0.76} & -0.63 \\
CoSQL  & \textbf{-0.71} & -0.64 \\
BIRD   & -0.35 & -0.37 \\
\bottomrule
\end{tabular}
\end{table}



\section{Controlling for Alternative Explanations}
\label{sec:critical}

We consider four alternative hypotheses that could potentially
explain our findings. In evaluating each, we note that any
explanation must account not only for the accuracy decline
itself, but for its differential magnitude across benchmarks:
the steepest drops occur on the oldest benchmark (Spider),
while BIRD---despite being the most structurally complex of
the four---shows minimal degradation. Generic
paraphrase-invariance or compositionality limitations would
predict similar decay patterns across datasets, which is not
what we observe.

\begin{figure}
    \centering
    \includegraphics[width=\linewidth]{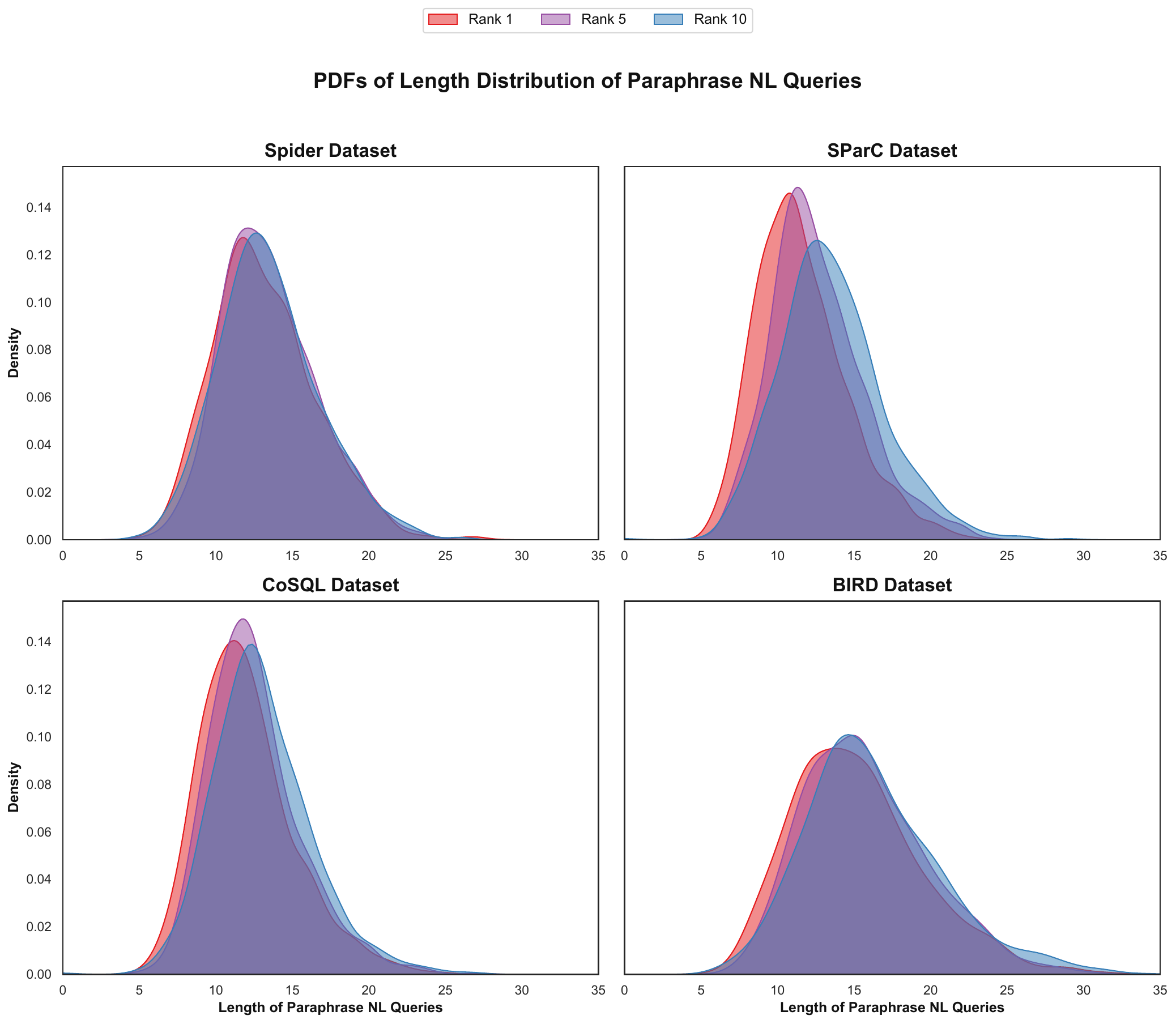}
    \caption{Probability Density Functions (PDFs) of Paraphrase Length Distributions for Different Ranks in Multiple Datasets. The plot compares the length distributions of paraphrased natural language queries (NLQs) across four datasets: Spider, SPARQL, CoSQL, and BIRD. The colors represent different ranks, with Rank 1 (red) corresponding to paraphrased queries with the closest syntactic similarity to the original query, Rank 10 (blue) representing paraphrased queries with the most distinct syntactic differences, and Rank 5 (purple) representing queries with moderate syntactic similarity. }
    \label{fig:length_distribution}
\end{figure}

First, we test whether the observed degradation in execution accuracy with increasing syntactic distance is driven by the \emph{length} of the generated queries rather than by their syntactic structure. 
In particular, one might suspect that higher-ranked paraphrases (i.e., those with greater syntactic divergence) lead to lower accuracy simply because they tend to be longer. 
To examine this, we plot the distribution of query lengths for syntactic ranks 1, 5, and 10 across the four benchmarks. 
As shown in Figure~\ref{fig:length_distribution}, the distributions are broadly similar across ranks, indicating that the performance decline in older benchmarks such as Spider is not an artifact of query length.

\begin{figure}
    \centering    
    \includegraphics[width=1\linewidth]{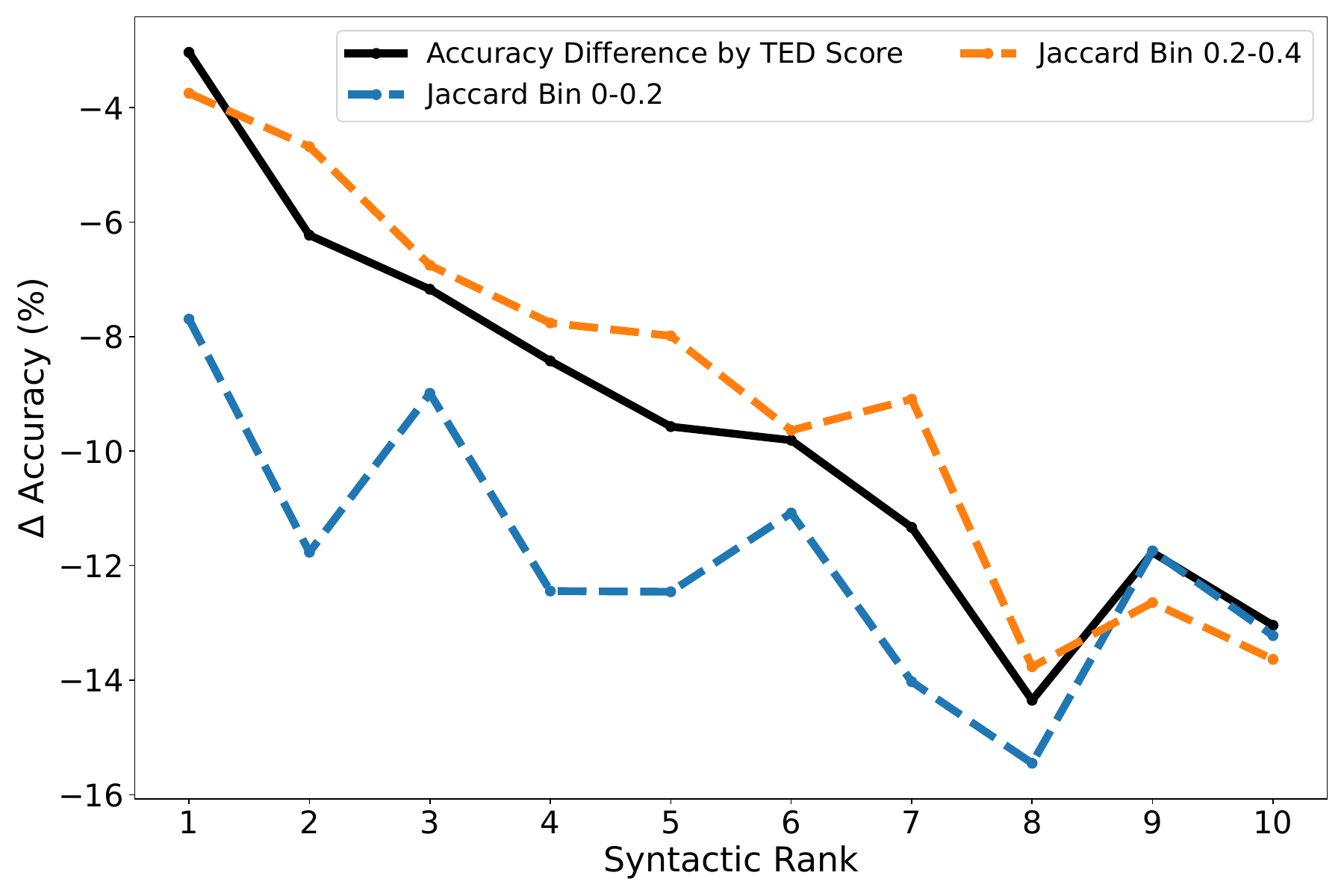}
    \caption{
Accuracy as a function of paraphrase rank for GPT-4o-mini, conditioned on Jaccard similarity. 
The black curve shows the unfiltered accuracy--rank trend, while the colored curves correspond to two bins of Jaccard-overlap between the original and paraphrased queries. 
The consistent downward slope across both bins indicates that SPENCE primarily measures syntactic robustness rather than surface-form similarity.
}
    \label{fig:lexical_stratified}
\vspace{-10pt}
\end{figure}

We further examine whether the observed rank sensitivity might be confounded by lexical overlap between the original and paraphrased queries. 
To test this, we perform a stratified \emph{apples-to-apples} analysis based on the Jaccard lexical-overlap score between each paraphrase and its corresponding original query, defined as 
$J(A,B) = \tfrac{|A \cap B|}{|A \cup B|}$, 
where $A$ and $B$ denote the sets of tokens in the original and paraphrased queries, respectively. 
Paraphrases are grouped into two overlap bins (0.0--0.2, 0.2--0.4), and within each bin we compute Kendall’s~$\tau$ between execution accuracy and syntactic rank. 
If the degradation were purely lexical, accuracy should remain stable once overlap is controlled. 
However, we find that the negative correlation with syntactic rank persists within all overlap strata, indicating that the trend is genuinely driven by structural, rather than lexical, divergence. 
Figure~\ref{fig:lexical_stratified} illustrates this effect for GPT-4o-mini: the unfiltered accuracy--rank curve (black) is shown alongside curves conditioned on the two Jaccard-overlap bins (colored lines). 
The consistent downward slope across bins confirms that SPENCE primarily probes syntactic robustness rather than surface-form similarity.

The distribution of Jaccard for all the benchmarks for a few of the ranks is provided in Appendix \ref{app:jaccard} Figure \ref{fig:jaccard_pdfs}. 

Lastly, we ask whether the stronger rank sensitivity observed in Spider may instead reflect \emph{schema-linking dependence}. 
If Spider queries rely more heavily on lexical and structural alignment between the natural-language question and the database schema, syntactic paraphrasing could disrupt these links\footnote{For example, consider the case that the schema uses Table with name "Student" and the original query also refers to the number of students. Then, if in the new query we replace "student" with "learner", it might break the schema-link.}, leading to larger performance drops. 
In contrast, benchmarks like BIRD may exhibit weaker schema-linking bias, and thus show less degradation under syntactic perturbation. 

\begin{figure}
    \centering    \includegraphics[width=1\linewidth]{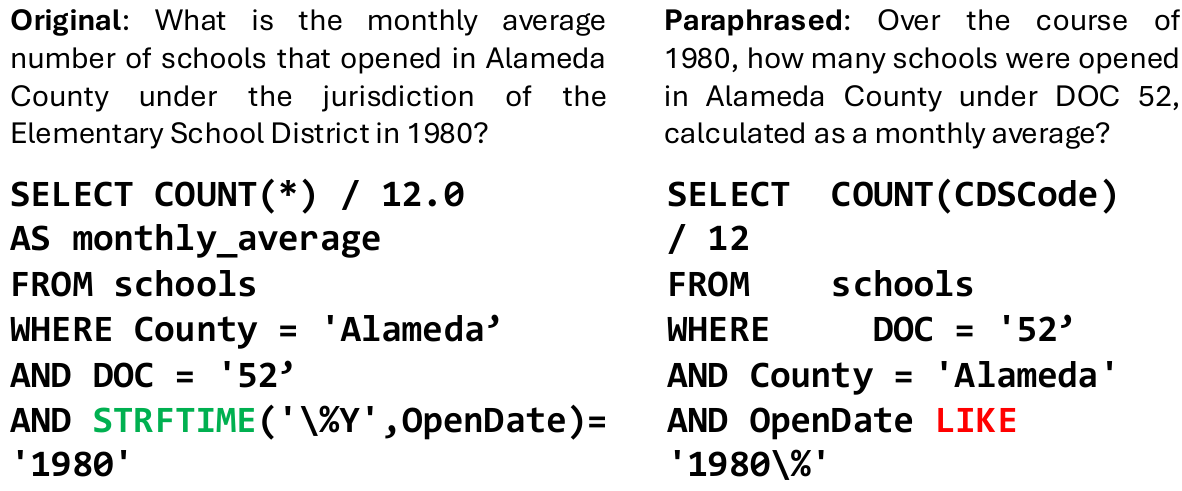}
    \includegraphics[width=1\linewidth]{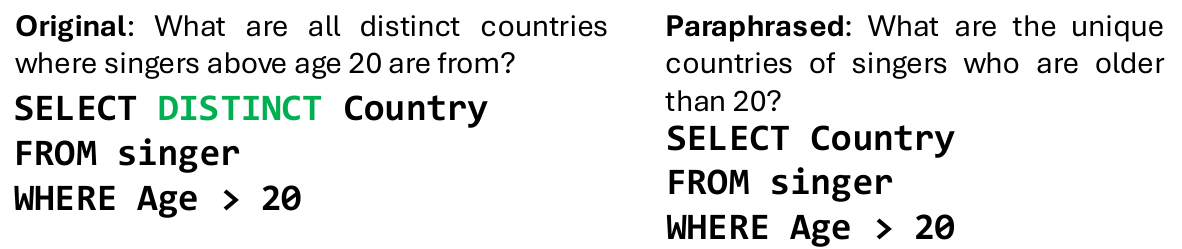}
    
    \caption{Samples of original and paraphrased queries for BIRD (top example) and Spider (bottom example), and the corresponding SQL queries. The paraphrased queries show no sign of schema linkage issue (like deviation from the table-names on the schema nor the column-names). Still, due to the semantical variations, the LLM is unable to generate a correct SQL query for the paraphrased queries. In the example from BIRD, it uses \textcolor{red}{LIKE} instead of filtering via \textcolor{green}{STRFTIME}. Note that, \textcolor{red}{LIKE} might not work as expected if the date is stored in a different format, or if the year is not always at the beginning of the string. In the example from Spider, apparently it missed to add \textcolor{green}{DISTINCT} to avoid repetition of countries.}
    \label{fig:bird_spider_examples}
\end{figure}

For a given LLM, we examined several cases where the original natural language (NL) query resulted in a successful execution, whereas its paraphrased counterpart failed. 
Among the inspected samples, we did not observe any instances of broken schema linking. 
Figure~\ref{fig:bird_spider_examples} provides one example each from the BIRD and Spider datasets, illustrating the original and paraphrased NL queries alongside their corresponding SQL queries. 
A closer inspection reveals that all key lexical anchors used for schema linking are preserved. 
For instance, in the BIRD example, the paraphrased NL query retains the keywords \texttt{"schools"}, \texttt{"County"}, and \texttt{"open date"}, which align respectively with \texttt{table=schools}, \texttt{column=County}, and \texttt{column=OpenDate} in the schema. 
Additionally, the identifier \texttt{DOC=52} is drawn from the \texttt{Evidence} section of the schema and correctly incorporates the filter values \texttt{"Alameda County"} and \texttt{"1980"}. 
Hence, no information loss or linking error occurs. 
Similarly, in the Spider example, the words \texttt{"singer"} and \texttt{"country"} correctly map to \texttt{table=singer} and \texttt{column=Country} in the schema (see Appendix~\ref{appndix:schema_linking_check} for the schema details).

\begin{figure}[t]
  \centering
  \includegraphics[width=\linewidth]{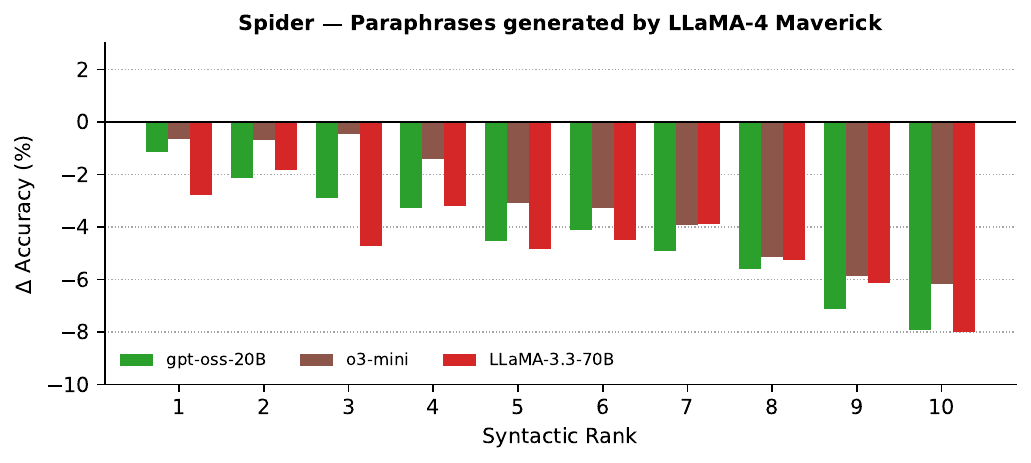}
  \caption{Difference in execution accuracy (Paraphrase~--~Original) on
  Spider when paraphrases are generated by LLaMA-4 Maverick instead of
  GPT-5, for three downstream NL2SQL models. Bars below the zero line
  indicate performance degradation. The monotonic decline with syntactic
  rank and the slope magnitudes mirror those observed under GPT-5
  paraphrasing,
  indicating that the SPENCE signal is not an artifact of the paraphrase
  generator.}
  \label{fig:maverick-control}
\end{figure}

Finally, we examine whether the observed rank sensitivity could be an artifact of the paraphrase generator itself, rather than a property of the underlying benchmarks. Because all paraphrases in our main experiments are produced by GPT-5, one might suspect that the accuracy decline reflects
GPT-5's stylistic idiosyncrasies, or potential exposure of GPT-5 to these benchmarks during its own pretraining. If either were the case, substituting
a different paraphrase generator should materially change the accuracy--rank curves. To test this, we repeat the paraphrase generation stage on Spider using
LLaMA-4 Maverick in place of GPT-5, holding the remainder of the SPENCE pipeline fixed: schema conditioning, TED-based ranking, embedding-based
filtering, and downstream NL2SQL evaluation. Figure~\ref{fig:maverick-control} reports the resulting $\Delta$ Accuracy per paraphrase rank for three downstream models spanning open-source and reasoning families. The monotonic decline with syntactic rank and the slope magnitudes are
comparable to those obtained under GPT-5 paraphrasing: gpt-oss-20B drops from 77.05 on the original queries to 69.11 at rank~10, LLaMA-3.3-70B from 76.41 to 68.42, and o3-mini from 76.37 to 70.20. The negative correlation with syntactic rank therefore persists across paraphrase generators, indicating that the SPENCE signal is robust to paraphrase-generator choice and is unlikely to be explained solely by GPT-5-specific stylistic or exposure effects.

\section{Conclusions}
\label{sec:conclusion}

We introduced \textbf{SPENCE} (\emph{Syntactic Probing and Evaluation of NL2SQL Contamination Effects}), 
a controlled syntactic probing framework for detecting and quantifying contamination in natural language to SQL (NL2SQL) benchmarks. 
By generating ranked sets of syntactically diverse paraphrases and measuring execution accuracy as a function of structural distance, SPENCE offers a reproducible and model-agnostic method for distinguishing genuine generalization from memorization of benchmark artifacts. 
Our experiments across four NL2SQL benchmarks reveal consistent rank-sensitivity on older datasets—Spider, SParC, and CoSQL—but not on the more recent BIRD benchmark, suggesting that contamination-like rank sensitivity is more pronounced in older benchmarks than in more recent ones.
These results underscore the need for syntactic-probing evaluations as a critical complement to standard leaderboard reporting. 
In future work, we plan to extend SPENCE beyond the NL2SQL domain to text-to-code and question answering tasks, and to incorporate automated paraphrase quality controls to support large-scale, systematic robustness auditing of language models.

\section{Limitations}
\label{sec:limitations}

Our analysis is limited to four public NL2SQL benchmarks (Spider, SParC, CoSQL, and BIRD), which may not fully represent the linguistic or schema diversity of real-world databases. 
The paraphrases are generated automatically by large language models and filtered using syntactic metrics (TED, cosine similarity), which could introduce generation biases. 
Moreover, we focus on execution accuracy as the primary metric; while informative, it conflates multiple error sources such as schema linking and logical form generation. In the multi-turn settings (SParC, CoSQL), we restrict paraphrasing to the final user turn; extending SPENCE to perturb earlier turns is possible, but would require additional consistency constraints over the dialogue context, and we leave this to future work. Our framework focuses on the syntactic axis of contamination;
extending SPENCE to sample-level probes (e.g., generating new but
semantically equivalent NL--SQL pairs in the style of ConStat) and
to semantic contamination is a natural direction for future work. Our evaluation also focuses on general-purpose LLMs rather than SQL-specialized systems such as OmniSQL or Arctic-Text2SQL-R1; extending SPENCE to such models is an important direction for future work. More broadly, the temporal gradient we observe is correlational: older benchmarks may differ from newer ones not only in prior exposure risk, but also in difficulty, annotation style, or distributional properties.
Finally, our use of Kendall’s~$\tau$ captures monotonic sensitivity trends but does not establish causal explanations for accuracy degradation under syntactic variation.

\bibliography{iclr2024_conference}

\appendix

\section{TED Computation}
\label{app:ted}
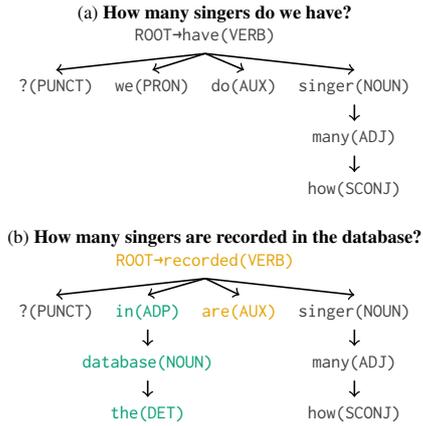
\begin{figure}[t]
\centering
\scriptsize

(a) \textbf{How many singers do we have?}\\[-2pt]
\begin{forest} compact
[{\textcolor{match}{ROOT→have(VERB)}} 
  [{\textcolor{match}{singer(NOUN)}} 
    [{\textcolor{match}{many(ADJ)}} 
      [{\textcolor{match}{how(SCONJ)}}]
    ]
  ]
  [{\textcolor{match}{do(AUX)}}]
  [{\textcolor{match}{we(PRON)}}]
  [{\textcolor{match}{?(PUNCT)}}]
]
\end{forest}

\vspace{6pt}

(b) \textbf{How many singers are recorded in the database?}\\[-2pt]
\begin{forest} compact
[{\textcolor{update}{ROOT→recorded(VERB)}} 
  [{\textcolor{match}{singer(NOUN)}} 
    [{\textcolor{match}{many(ADJ)}} 
      [{\textcolor{match}{how(SCONJ)}}]
    ]
  ]
  [{\textcolor{update}{are(AUX)}}]
  [{\textcolor{insert}{in(ADP)}} 
    [{\textcolor{insert}{database(NOUN)}} 
      [{\textcolor{insert}{the(DET)}}]
    ]
  ]
  [{\textcolor{match}{?(PUNCT)}}]
]
\end{forest}

\vspace{-4pt}
\caption{Dependency-tree alignment for two semantically equivalent questions: 
(a) ``How many singers do we have?'' and (b) ``How many singers are recorded in the database?''. 
Each node denotes a token annotated with its Universal Part-of-Speech (POS) tag, 
and edges indicate head--dependent relations. 
Colors correspond to minimal tree-edit operations: 
\textcolor{match}{gray} = \emph{match} (preserved structure), 
\textcolor{update}{orange} = \emph{update} (morphological substitution, 
e.g., \emph{do}$\rightarrow$\emph{are}, \emph{have}$\rightarrow$\emph{recorded}), and 
\textcolor{insert}{teal} = \emph{insert} (new constituents such as \emph{in the database}). 
POS tags follow the Universal Dependencies standard: 
\texttt{VERB} (verb), \texttt{NOUN} (noun), \texttt{ADJ} (adjective), 
\texttt{SCONJ} (subordinating conjunction), \texttt{AUX} (auxiliary verb), 
\texttt{ADP} (adposition/preposition), \texttt{DET} (determiner), 
\texttt{PRON} (pronoun), and \texttt{PUNCT} (punctuation). 
The shared interrogative phrase ``how many singers'' remains identical across both trees, 
while the predicate transitions from an active possessive to a passive recorded-state form.}
\label{fig:tree_edit_vertical}
\end{figure}

To quantify syntactic divergence between original and paraphrased queries, 
we compute the Tree Edit Distance (TED) between their dependency parses. 
Figure~\ref{fig:tree_edit_vertical} illustrates the alignment between two semantically equivalent questions drawn from our paraphrase set. 
TED measures the minimal number of structural operations—insertions, deletions, and substitutions—required to transform one dependency tree into another. 
In this example, the interrogative span (\emph{how many singers}) remains structurally identical across both trees, whereas the predicate transitions from an active possessive form (\emph{have}) to a passive recorded-state construction (\emph{are recorded in the database}). 
Such substitutions correspond to \emph{update} operations, while the addition of prepositional modifiers (e.g., \emph{in the database}) contributes \emph{insert} operations. 
The cumulative cost of these edits defines the syntactic distance rank used throughout SPENCE to order paraphrases and correlate model performance with structural divergence.

\section{Prompt Template}
\label{app:prompt}
\begin{lstlisting}
Given the following database schema and an SQL query, generate {num_queries} distinct natural language questions that describe the purpose and output of the SQL query.

{schema_definitions}

SQL Query:
{sql_query}

Instructions:
1. Generate {num_queries} natural language questions that reflect the intent of the SQL query.
2. Each question should vary in phrasing, structure, and wording, but all questions must remain logically equivalent.
3. Do not include explanations, task descriptions, or any additional comments in the output.

Output Format:
1. <First question>
2. <Second question>
...
{num_queries}. <Nth question>.
\end{lstlisting}

\section{Paraphrase Examples}
\label{app:para_example}

Table \ref{tab:ted-71-ruled} presents examples of the paraphrased queries sorted by their TED distance from BIRD dev-set benchmark. 
\begin{table*}[t]
\centering
\caption{Examples of paraphrased queries for one original question from the BIRD dataset, sorted by normalized Tree Edit Distance (TED) from the original. Lower TED indicates closer syntactic similarity.}
\label{tab:ted-71-ruled}

\begin{minipage}{\linewidth}
\textbf{Original query:} \emph{How many active and closed District Community Day Schools are there in the county of Alpine?}
\end{minipage}\vspace{0.5em}

\begin{tabularx}{\linewidth}{
  >{\raggedright\arraybackslash}X
  >{\raggedleft\arraybackslash}p{1.6cm}
}
\toprule
\textbf{Paraphrase query} & \textbf{TED (norm)} \\
\midrule
How many schools in Alpine County are either closed or currently active? & 0.4375 \\
How many schools in Alpine County are classified as either active or closed? & 0.4545 \\
How many schools in the county of Alpine have a status of closed or active? & 0.4571 \\
Can you determine how many schools in Alpine County are listed as either active or closed? & 0.5000 \\
Can you count the schools in Alpine County that are marked as either active or closed? & 0.5556 \\
What is the total number of schools in Alpine County with a status of "Closed" or "Active"? & 0.5610 \\
What is the total count of schools in Alpine County with their status listed as "Closed" or "Active"? & 0.5714 \\
What is the number of schools in Alpine County that are either operational or were previously closed? & 0.5946 \\
What is the number of schools in Alpine County that have an "Active" or "Closed" status? & 0.6000 \\
Provide the count of schools within Alpine County that are currently active or have been closed. & 0.6389 \\
\bottomrule
\end{tabularx}
\end{table*}

\section{Effect of LLM Temperature}
\label{app:temperature}

\begin{figure}
    \centering
    \includegraphics[width=.9\linewidth]{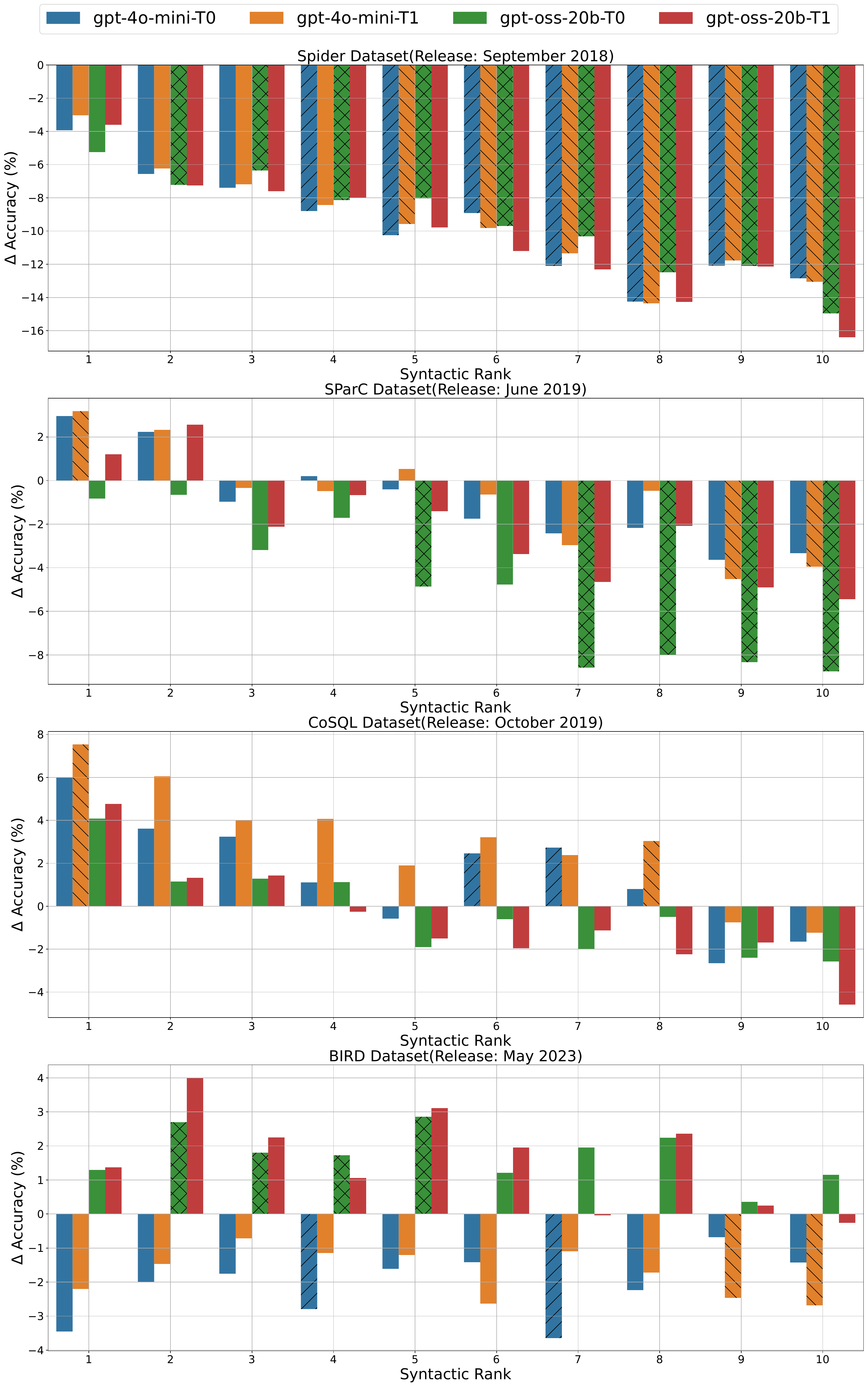}
    \caption{Difference in execution accuracy (Paraphrase\,–\,Original queries) for GPT-OSS-20B and GPT4o-mini across paraphrase ranks on four NL2SQL benchmarks — Spider (Sep 2018), SParC (Jun 2019), CoSQL (Oct 2019), and BIRD (May 2023) — shown in release order.
Similar to Figure~\ref{fig:accuracy-by-rank}, anything above the zero line indicate paraphrased queries outperformed the original queries and vice versa. We show the temperature=0 and temperature=1 with 'T0' and 'T1', respectively, added at the end of LLM names.}
    \label{fig:temperature_effect}
\end{figure}

We wanted to see how robust the results are when the temperature of the LLM  changes. So, we ran the experiments with temperature=0, for the LLM for both the query generation and SQL generation. We tried GPT-OSS-20B and GPT4o-mini\cite{openai2024gpt} for all datasets, and got the performance drop across all ranks. 
Figure~\ref{fig:temperature_effect} shows the corresponding results, where we present the performance drop for both temperatures 0.0 and 1.0, represented via 'T0' and 'T1' on the LLM names. As shown, the LLM temperature does not change the trend of the results, still the same trend on the performance-drop with increasing the rank exist. So, the temperature does not affect the leakage and with a wide range of temperatures we still observe the same issue. 

\section{Full Table for Kendall values}
\label{app:kendall}
Detailed version of our results is presented in Table \ref{tab:kendall-combined}.
\begin{table}
\centering
\caption{Kendall’s~$\tau$ between paraphrase rank and $\Delta$ Accuracy (paraphrase accuracy minus original accuracy) for each model and benchmark. Two columns per model: ``All ranks'' and ``Ranks~$\geq 3$''. Values in \textbf{bold} have 95\% bootstrap confidence intervals entirely below zero, indicating statistically significant rank sensitivity at $\alpha=0.05$.}
\label{tab:kendall-combined}
\begin{adjustbox}{width=.5\textwidth}
\begin{tabular}{l @{\hspace{0.2cm}} l cc cc}
\toprule
Dataset & ~Model & $\tau$ All ranks & 95\% CI & $\tau$ Ranks $\geq3$ & 95\% CI \\
\midrule
Spider & GPT-OSS-120 & \textbf{-0.78} & [-1.00,\,-0.46] & \textbf{-0.79} & [-1.00,\,-0.36] \\
Spider & GPT-OSS-20  & \textbf{-0.91} & [-1.00,\,-0.76] & \textbf{-0.86} & [-1.00,\,-0.34] \\
Spider & GPT-4o-mini     & \textbf{-0.91} & [-1.00,\,-0.35] & \textbf{-0.86} & [-1.00,\,-0.36] \\
Spider & LLaMA3.3-70B      & \textbf{-0.87} & [-1.00,\,-0.62] & \textbf{-0.86} & [-1.00,\,-0.51] \\
Spider & LLaMA3.1-405B   & \textbf{-0.96} & [-1.00,\,-0.74] & \textbf{-1.00} & [-1.00,\,-1.00] \\
Spider & O3-mini   & \textbf{-0.91} & [-1.00,\,-0.65] & \textbf{-0.93} & [-1.00,\,-0.74] \\
\midrule
SParC  & GPT-OSS-120 & \textbf{-0.82} & [-0.95,\,-0.67] & \textbf{-0.79} & [-1.00,\,-0.45] \\
SParC  & GPT-OSS-20  & \textbf{-0.73} & [-1.00,\,-0.32] & -0.64 & [-1.00,\,-0.08] \\
SParC  & GPT-4o-mini     & \textbf{-0.73} & [-1.00,\,-0.38] & -0.57 & [-1.00,\,-0.13] \\
SParC  & LLaMA3.3-70B      & \textbf{-0.87} & [-1.00,\,-0.53] & \textbf{-0.79} & [-1.00,\,-0.23] \\
SParC  & LLaMA3.1-405B   & -0.64 & [-0.92,\,-0.30] & -0.43 & [-0.83,\,0.09] \\
SParC  & O3-mini  & -0.69 & [-0.95,\,-0.30] & -0.57 & [-1.00,\,-0.16] \\
\midrule
CoSQL~ & GPT-OSS-120 & \textbf{-0.73} & [-1.00,\,-0.25] & -0.64 & [-1.00,\,-0.01] \\
CoSQL  & GPT-OSS-20  & \textbf{-0.78} & [-1.00,\,-0.40] & -0.71 & [-1.00,\,0.04] \\
CoSQL  & GPT-4o-mini     & \textbf{-0.78} & [-1.00,\,-0.36] & -0.64 & [-1.00,\,-0.06] \\
CoSQL  & LLaMA3.3-70B      & \textbf{-0.78} & [-1.00,\,-0.52] & \textbf{-0.79} & [-1.00,\,-0.29] \\
CoSQL  & LLaMA3.1-405B   & -0.64 & [-0.83,\,-0.40] & \textbf{-0.57} & [-0.91,\,-0.21] \\
CoSQL  & O3-mini         & -0.69 & [-1.00,\,-0.16] & -0.50 & [-1.00,\,0.32] \\
\midrule
BIRD   & GPT-OSS-120 & -0.47 & [-0.85,\,-0.02] & -0.50 & [-1.00,\,0.16] \\
BIRD   & GPT-OSS-20  & -0.42 & [-0.87,\,0.12]  & -0.43 & [-0.84,\,0.14] \\
BIRD   & GPT-4o-mini     & -0.33 & [-0.82,\,0.53]  & -0.64 & [-1.00,\,0.09] \\
BIRD   & LLaMA3.3-70B      & -0.11 & [-0.73,\,0.32]  & -0.14 & [-0.77,\,1.00] \\
BIRD   & LLaMA3.1-405B   & -0.39 & [-0.88,\,0.21]  & -0.24 & [-1.00,\,0.62] \\
BIRD   & O3-mini         & -0.38 & [-0.88,\,0.05]  & -0.36 & [-1.00,\,0.25] \\
\bottomrule
\end{tabular}
\end{adjustbox}
\end{table}

\section{Length Distribution of Paraphrase NL Queries}

As shown in Figure~\ref{fig:length_distribution}, the probability density functions (PDFs) of paraphrased query lengths are largely similar across ranks and datasets. 
Although higher-rank paraphrases (e.g., Rank~10 in blue) tend to exhibit slightly broader or right-shifted distributions compared to lower-rank paraphrases (e.g., Rank~1 in red), the overall differences are minute. 
This indicates that the observed decline in execution accuracy with increasing syntactic rank (Section 5) cannot be attributed simply to differences in query length.


\section{Jaccard similarity Distribution of Paraphrase NL Queries}
\label{app:jaccard}

\begin{figure}
    \centering    
        \includegraphics[width=1\linewidth]{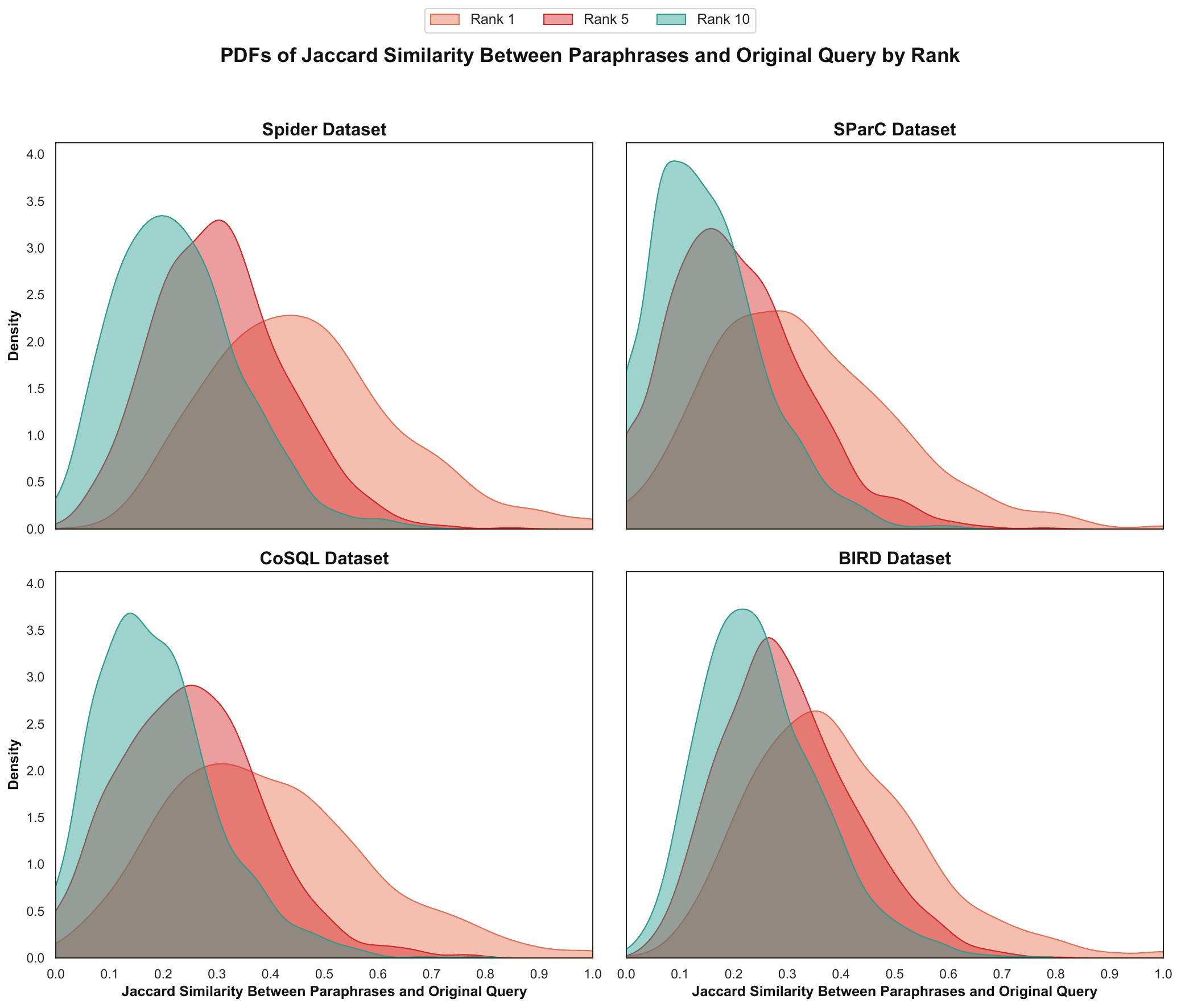}
    \caption{Probability Density Functions (PDFs) of Jaccard Similarity Between Paraphrases and Original Query for Different Ranks in Multiple Datasets. The plot compares the Jaccard similarity distributions of paraphrased natural language queries (NLQs) across four datasets: Spider, SPARQL, CoSQL, and BIRD. The colors represent different ranks, with Rank 1 (light red) corresponding to paraphrased queries with the closest similarity to the original query, Rank 10 (cyan) representing paraphrased queries with the most distinct differences in similarity, and Rank 5 (dark red) representing queries with moderate similarity to the original.}
    \label{fig:jaccard_pdfs}
\end{figure}

Figure~\ref{fig:jaccard_pdfs} shows the probability density functions (PDFs) of Jaccard similarity between paraphrased and original natural language queries across different ranks and datasets. 
As expected, Rank~1 paraphrases (light red) exhibit the highest Jaccard similarity, closely matching the lexical content of the original query, while higher ranks (e.g., Rank~10 in cyan) show progressively lower overlap. 
However, the overall separation between ranks remains moderate, suggesting that even paraphrases with substantial syntactic divergence retain partial lexical overlap with the source query. 
This pattern indicates that the SPENCE ranking procedure successfully produces a graded spectrum of paraphrases—ranging from near-identical reformulations to substantially reworded variants—without collapsing into either trivial copies or semantically unaligned outputs.

\section{Schema Linking Preservation}\label{appndix:schema_linking_check}

Tables~\ref{tb:bird_examples} and \ref{tb:spider_examples} show some examples of for Bird and Spider datasets where the paraphrased queries keep the schema linking, while LLMs fail to get the correct SQL query based on them. On each dataset, there are several rows where {\bf Original} and {\bf Paraphrased-Rx} show the original or paraphrased samples, in which "Rx" refers to the sample with Rank$=x \in \{1,\dots, 10\}$. 
Also, Listings \ref{appndix:schema_bird} and \ref{appndix:schema_spider} shows the corresponding schema. The example for some ranks are missing, since the generated query did not satisfy the bare minimum quality bars on the query generation. This happens more for smaller queries, like the one in Table~\ref{tb:spider_examples}, and less often for longer more complex queries, like the one in Table~\ref{tb:bird_examples}.

As shown in both tables, the generated queries hold all the required information to create a SQL query, while keeping all the keywords aligned with the schema. This is the result of our design choice to pass both the gold-query and the schema for generating new queries, so that there is no misalignment and schema-linking issue between the schema and the new queries. 

\section{Human Validation of Semantic Equivalence}
\label{app:human-eval}
 
To verify that TED-based ranking reflects structural rather than semantic
divergence, we conduct a small human study on paraphrase fidelity. We sample 100 base questions from the Spider development set and, for
each, extract the paraphrases at ranks 1, 5, and 10 produced by our pipeline (Section~\ref{sec:methods})—yielding 300 paraphrase--original
pairs in total. Annotators were shown each pair without rank information and asked to assign one of three labels: \emph{equivalent} (the paraphrase
preserves the meaning of the original question), \emph{partially equivalent} (the paraphrase shifts or omits a non-trivial part of the original meaning),
or \emph{not equivalent} (the paraphrase asks a meaningfully different question). The rank at which each paraphrase was drawn was hidden from
annotators to avoid anchoring effects.
 Across the 300 pairs, more than 85\% are judged semantically equivalent
to their originals, and the proportion of equivalent judgments does not
vary systematically with rank: rank-10 paraphrases are judged faithful
at a rate comparable to rank-1 paraphrases. The absence of a rank trend in semantic judgments is consistent with our interpretation that TED
ranks capture structural reorganization (clause reordering, predicate substitution, passive/active shifts) rather than changes in underlying
meaning. This supports the use of TED-based ranking as a probe for syntactic—rather than semantic—robustness, and ensures that the accuracy decline observed at higher ranks is not an artifact of paraphrase quality.
 
We note that this is a small-scale validation intended to confirm that
the filtering pipeline (Section~\ref{sec:methods}) produces
meaning-preserving paraphrases across rank tiers; a larger annotation
effort with inter-annotator agreement statistics would be a natural
extension, and we leave this to future work.

\begin{table*}
\centering
\caption{Samples of original and paraphrased queries and the corresponding SQL queries, for a sample from Bird dataset. The paraphrased queries show no sign of schema linkage issue, and as a result of semantic edits, the LLM is unable to generate a correct SQL query. On each row, Original/Paraphrased-Rx shows if it is the original or paraphrased samples, and "Rx" refers to the sample with Rank$=x \in \{1,\dots, 10\}$.}
\label{tb:bird_examples}
\begin{tabular}{p{7.9cm} @{\hspace{0.2cm}} p{7.9cm}} \toprule
Query  & SQL  \\ \hline

{\bf Original}: What is the monthly average number of schools that opened in Alameda County under the jurisdiction of the Elementary School District in 1980? & {\tt SELECT COUNT(*) / 12.0 AS monthly\_average FROM schools WHERE County = 'Alameda' AND DOC = '52'  AND STRFTIME('\%Y', OpenDate) = '1980'} \\ \hline

{\bf Paraphrased-R1}: What is the monthly average of schools that opened in Alameda County in 1980 with a DOC value of 52?  & {\tt SELECT      COUNT(DISTINCT School)  FROM      schools  WHERE      SUBSTR(OpenDate, 1, 4) = '1980'      AND County = 'Alameda'      AND DOC = '52'} \\ \hline

{\bf Paraphrased-R3}: For Alameda County in 1980, what is the monthly average of schools that opened and had a DOC value equal to 52?  & {\tt SELECT      COUNT(CDSCode)  FROM      schools  WHERE      County = 'Alameda'      AND DOC = '52'      AND STRFTIME('\%Y', OpenDate) = '1980'} \\ \hline

{\bf Paraphrased-R4}: What is the calculated monthly average of new schools in Alameda County during 1980 with a DOC value of 52?  & {\tt SELECT      AVG(CASE          WHEN STRFTIME('\%Y', OpenDate) = '1980'          THEN 1          ELSE 0      END) / 12 AS monthly\_average\_new\_schools FROM schools WHERE County = 'Alameda' AND DOC = '52'} \\ \hline

{\bf Paraphrased-R5}: What is the average number of schools opened each month in Alameda County in 1980 under Department of Education code 52 & {\tt SELECT AVG(OpenDate)  FROM schools  WHERE County = 'Alameda' AND DOC = '52' AND CAST(STRFTIME('\%Y', OpenDate) AS INTEGER) = 1980;} \\ \hline

{\bf Paraphrased-R6}: What is the average number of schools opened per month in Alameda County in 1980 with a DOC value of 52? & {\tt SELECT      COUNT(DISTINCT School)  FROM      schools  WHERE      County = 'Alameda'  AND      SUBSTR(OpenDate, 1, 4) = '1980'  AND      DOC = '52'} \\ \hline

{\bf Paraphrased-R7}: How many schools, on average, opened each month in Alameda County in 1980 with the Department of Education code set to 52? & {\tt SELECT AVG(TotalSchools)  FROM (   SELECT      STRFTIME('\%Y-\%m', OpenDate) AS Month,      COUNT(CDSCode) AS TotalSchools   FROM      schools   WHERE      County = 'Alameda'     AND DOC = '52'     AND OpenDate BETWEEN '1980-01-01' AND '1980-12-31'   GROUP BY      STRFTIME('\%Y-\%m', OpenDate) );}\\ \hline

{\bf Paraphrased-R8}: How many schools were opened in Alameda County under the Department of Education code 52 during the year 1980, on average per month? & {\tt SELECT      COUNT(CDSCode) / 12.0 AS AverageSchoolsPerMonth FROM      schools WHERE      DOC = '52'      AND STRFTIME('\%Y', OpenDate) = '1980'} \\ \hline

{\bf Paraphrased-R9}: Over the course of 1980, how many schools were opened in Alameda County under DOC 52, calculated as a monthly average?                     & {\tt SELECT  COUNT(CDSCode) / 12 FROM    schools WHERE     DOC = '52'    AND County = 'Alameda'    AND OpenDate LIKE '1980\%'} \\ \hline

{\bf Paraphrased-R10}: On a monthly basis, how many schools were opened in the year 1980 in Alameda County under Department code 52? & {\tt SELECT      COUNT(DISTINCT School)  FROM      schools  WHERE      StatusType = 'Active'      AND County = 'Alameda'      AND OpenDate LIKE '1980-\%'     AND DOCType = '52'} \\ 
\bottomrule

\end{tabular}
\end{table*}

\begin{table*}
\centering
\caption{Samples of original and paraphrased queries and the corresponding SQL queries, for a sample from Spider dataset. Similar to Bird dataset, there is sign of schema linkage issue in the paraphrased queries, and due to semantic edits the LLM cannot generate a correct SQL query. }
\label{tb:spider_examples}
\begin{tabular}{p{7.9cm} @{\hspace{0.2cm}} p{7.9cm}} \toprule
query  & SQL  \\ \hline

{\bf Original}: What are all distinct countries where singers above age 20 are from?  & {\tt SELECT DISTINCT Country FROM singer WHERE Age \textgreater 20} \\ \hline

{\bf Paraphrased-R3}: What are the unique countries of singers who are older than 20?   & {\tt SELECT Country  FROM singer  WHERE Age > 20 }  \\ \hline
{\bf Paraphrased-R4}: Can you list the different countries represented by singers with an age greater than 20?   & {\tt SELECT Country  FROM singer  WHERE Age > 20}  \\ \hline
{\bf Paraphrased-R8}:  From how many unique countries do singers with an age greater than 20 hail?  & {\tt SELECT COUNT(DISTINCT Country)  FROM singer  WHERE Age > 20}  \\ \hline
{\bf Paraphrased-R9}: Which unique countries are associated with singers aged over 20?   & {\tt SELECT Country  FROM singer  WHERE Age > 20}  \\ \hline
\end{tabular}
\end{table*}

\clearpage
\onecolumn
\begin{center}
    
\begin{adjustbox}{width=5\textwidth}
\centering
\begin{lstlisting}[caption={SQL schema for the example of Bird dataset.}, label={appndix:schema_bird}]
CREATE TABLE schools (
	"CDSCode" TEXT NOT NULL,  "NCESDist" TEXT,  
    "NCESSchool" TEXT,  "StatusType" TEXT NOT NULL, "County" TEXT NOT NULL,  "District" TEXT NOT NULL, "School" TEXT, "Street" TEXT, "StreetAbr" TEXT, "City" TEXT, "Zip" TEXT, "State" TEXT, "MailStreet" TEXT, "MailStrAbr" TEXT, "MailCity" TEXT,  "MailZip" TEXT, "MailState" TEXT, "Phone" TEXT, "Ext" TEXT, "Website" TEXT, "OpenDate" DATE, "ClosedDate" DATE, "Charter" INTEGER, "CharterNum" TEXT, "FundingType" TEXT, "DOC" TEXT NOT NULL, "DOCType" TEXT NOT NULL, "SOC" TEXT, "SOCType" TEXT, "EdOpsCode" TEXT, "EdOpsName" TEXT, "EILCode" TEXT, "EILName" TEXT, "GSoffered" TEXT, "GSserved" TEXT, "Virtual" TEXT, "Magnet" INTEGER, "Latitude" REAL, "Longitude" REAL, "AdmFName1" TEXT, "AdmLName1" TEXT, "AdmEmail1" TEXT, "AdmFName2" TEXT, "AdmLName2" TEXT, "AdmEmail2" TEXT, "AdmFName3" TEXT, "AdmLName3" TEXT, "AdmEmail3" TEXT, 
	"LastUpdate" DATE NOT NULL, 
	PRIMARY KEY ("CDSCode")
)
/*
3 rows from schools table:
CDSCode	NCESDist	NCESSchool	StatusType	County	District	School	Street	StreetAbr	City	Zip	State	MailStreet	MailStrAbr	MailCity	MailZip	MailState	Phone	Ext	Website	OpenDate	ClosedDate	Charter	CharterNum	FundingType	DOC	DOCType	SOC	SOCType	EdOpsCode	EdOpsName	EILCode	EILName	GSoffered	GSserved	Virtual	Magnet	Latitude	Longitude	AdmFName1	AdmLName1	AdmEmail1	AdmFName2	AdmLName2	AdmEmail2	AdmFName3	AdmLName3	AdmEmail3	LastUpdate
01100170000000	0691051	None	Active	Alameda	Alameda County Office of Education	None	313 West Winton Avenue	313 West Winton Ave.	Hayward	94544-1136	CA	313 West Winton Avenue	313 West Winton Ave.	Hayward	94544-1136	CA	(510) 887-0152	None	www.acoe.org	None	None	None	None	None	00	County Office of Education (COE)	None	None	None	None	None	None	None	None	None	None	37.658212	-122.09713	L Karen	Monroe	lkmonroe@acoe.org	None	None	None	None	None	None	2015-06-23
01100170109835	0691051	10546	Closed	Alameda	Alameda County Office of Education	FAME Public Charter	39899 Balentine Drive, Suite 335	39899 Balentine Dr., Ste. 335	Newark	94560-5359	CA	39899 Balentine Drive, Suite 335	39899 Balentine Dr., Ste. 335	Newark	94560-5359	CA	None	None	None	2005-08-29	2015-07-31	1	0728	Directly funded	00	County Office of Education (COE)	65	K-12 Schools (Public)	TRAD	Traditional	ELEMHIGH	Elementary-High Combination	K-12	K-12	P	0	37.521436	-121.99391	None	None	None	None	None	None	None	None	None	2015-09-01
01100170112607	0691051	10947	Active	Alameda	Alameda County Office of Education	Envision Academy for Arts \& Technology	1515 Webster Street	1515 Webster St.	Oakland	94612-3355	CA	1515 Webster Street	1515 Webster St.	Oakland	94612	CA	(510) 596-8901	None	www.envisionacademy.org/	2006-08-28	None	1	0811	Directly funded	00	County Office of Education (COE)	66	High Schools (Public)	TRAD	Traditional	HS	High School	9-12	9-12	N	0	37.80452	-122.26815	Laura	Robell	laura@envisionacademy.org	None	None	None	None	None	None	2015-06-18
*/


CREATE TABLE frpm (
	"CDSCode" TEXT NOT NULL, "Academic Year" TEXT, "County Code" TEXT, "District Code" INTEGER, "School Code" TEXT, "County Name" TEXT, "District Name" TEXT, "School Name" TEXT, "District Type" TEXT, "School Type" TEXT, "Educational Option Type" TEXT, "NSLP Provision Status" TEXT, "Charter School (Y/N)" INTEGER, "Charter School Number" TEXT, "Charter Funding Type" TEXT, "IRC" INTEGER, "Low Grade" TEXT, "High Grade" TEXT, "Enrollment (K-12)" REAL, "Free Meal Count (K-12)" REAL, "Percent (%) Eligible Free (K-12)" REAL, "FRPM Count (K-12)" REAL, "Percent (%) Eligible FRPM (K-12)" REAL, "Enrollment (Ages 5-17)" REAL, "Free Meal Count (Ages 5-17)" REAL, "Percent (%) Eligible Free (Ages 5-17)" REAL, "FRPM Count (Ages 5-17)" REAL, "Percent (%) Eligible FRPM (Ages 5-17)" REAL, "2013-14 CALPADS Fall 1 Certification Status" INTEGER, 
    PRIMARY KEY ("CDSCode"), 
	FOREIGN KEY("CDSCode") REFERENCES schools ("CDSCode")
)
/*
3 rows from frpm table:
CDSCode	Academic Year	County Code	District Code	School Code	County Name	District Name	School Name	District Type	School Type	Educational Option Type	NSLP Provision Status	Charter School (Y/N)	Charter School Number	Charter Funding Type	IRC	Low Grade	High Grade	Enrollment (K-12)	Free Meal Count (K-12)	Percent (\%) Eligible Free (K-12)	FRPM Count (K-12)	Percent (\%) Eligible FRPM (K-12)	Enrollment (Ages 5-17)	Free Meal Count (Ages 5-17)	Percent (\%) Eligible Free (Ages 5-17)	FRPM Count (Ages 5-17)	Percent (\%) Eligible FRPM (Ages 5-17)	2013-14 CALPADS Fall 1 Certification Status
01100170109835	2014-2015	01	10017	0109835	Alameda	Alameda County Office of Education	FAME Public Charter	County Office of Education (COE)	K-12 Schools (Public)	Traditional	None	1	0728	Directly funded	1	K	12	1087.0	565.0	0.519779208831647	715.0	0.657773689052438	1070.0	553.0	0.516822429906542	702.0	0.65607476635514	1
01100170112607	2014-2015	01	10017	0112607	Alameda	Alameda County Office of Education	Envision Academy for Arts \& Technology	County Office of Education (COE)	High Schools (Public)	Traditional	None	1	0811	Directly funded	1	9	12	395.0	186.0	0.470886075949367	186.0	0.470886075949367	376.0	182.0	0.484042553191489	182.0	0.484042553191489	1
01100170118489	2014-2015	01	10017	0118489	Alameda	Alameda County Office of Education	Aspire California College Preparatory Academy	County Office of Education (COE)	High Schools (Public)	Traditional	None	1	1049	Directly funded	1	9	12	244.0	134.0	0.549180327868853	175.0	0.717213114754098	230.0	128.0	0.556521739130435	168.0	0.730434782608696	1
*/


CREATE TABLE satscores (
	cds TEXT NOT NULL,   \# California Department Schools
	rtype TEXT NOT NULL, 
	sname TEXT,   \# school name
	dname TEXT,   \# district segment
	cname TEXT,   \# county name
	enroll12 INTEGER NOT NULL,   \# enrollment (1st-12nd grade)
	"NumTstTakr" INTEGER NOT NULL,  "AvgScrRead" INTEGER, "AvgScrMath" INTEGER, "AvgScrWrite" INTEGER, "NumGE1500" INTEGER, 
    PRIMARY KEY (cds), 
	FOREIGN KEY(cds) REFERENCES schools ("CDSCode")
)
/*
3 rows from satscores table:
cds	rtype	sname	dname	cname	enroll12	NumTstTakr	AvgScrRead	AvgScrMath	AvgScrWrite	NumGE1500
1100170000000	D	None	Alameda County Office of Education	Alameda	398	88	418	418	417	14
1100170109835	S	FAME Public Charter	Alameda County Office of Education	Alameda	62	17	503	546	505	9
1100170112607	S	Envision Academy for Arts \& Technology	Alameda County Office of Education	Alameda	75	71	397	387	395	5
*/
\end{lstlisting}
\end{adjustbox}
\end{center}

\clearpage
\onecolumn

\begin{adjustbox}{width=0.5\textwidth}
\centering
\begin{lstlisting}[caption={SQL schema for the example of Spider dataset.}, label={appndix:schema_spider}]
CREATE TABLE singer (
	"Singer_ID" INTEGER, 
	"Name" TEXT, 
	"Country" TEXT, 
	"Song_Name" TEXT, 
	"Song_release_year" TEXT, 
	"Age" INTEGER, 
	"Is_male" BOOLEAN, 
	PRIMARY KEY ("Singer_ID")
)
/*
3 rows from singer table:
Singer_ID	Name	Country	Song_Name	Song_release_year	Age	Is_male
1	Joe Sharp	Netherlands	You	1992	52	True
2	Timbaland	United States	Dangerous	2008	32	True
3	Justin Brown	France	Hey Oh	2013	29	True
*/


CREATE TABLE stadium (
	"Stadium_ID" INTEGER, 
	"Location" TEXT, 
	"Name" TEXT, 
	"Capacity" INTEGER, 
	"Highest" INTEGER, 
	"Lowest" INTEGER, 
	"Average" INTEGER, 
	PRIMARY KEY ("Stadium_ID")
)
/*
3 rows from stadium table:
Stadium_ID	Location	Name	Capacity	Highest	Lowest	Average
1	Raith Rovers	Stark's Park	10104	4812	1294	2106
2	Ayr United	Somerset Park	11998	2363	1057	1477
3	East Fife	Bayview Stadium	2000	1980	533	864
*/


CREATE TABLE concert (
	"concert_ID" INTEGER, 
	"concert_Name" TEXT, 
	"Theme" TEXT, 
	"Stadium_ID" TEXT, 
	"Year" TEXT, 
	PRIMARY KEY ("concert_ID"), 
	FOREIGN KEY("Stadium_ID") REFERENCES stadium ("Stadium_ID")
)
/*
3 rows from concert table:
concert_ID	concert_Name	Theme	Stadium_ID	Year
1	Auditions	Free choice	1	2014
2	Super bootcamp	Free choice 2	2	2014
3	Home Visits	Bleeding Love	2	2015
*/


CREATE TABLE singer_in_concert (
	"concert_ID" INTEGER, 
	"Singer_ID" TEXT, 
	PRIMARY KEY ("concert_ID", "Singer_ID"), 
	FOREIGN KEY("concert_ID") REFERENCES concert ("concert_ID"), 
	FOREIGN KEY("Singer_ID") REFERENCES singer ("Singer_ID")
)
/*
3 rows from singer_in_concert table:
concert_ID	Singer_ID
1	2
1	3
1	5
*/
\end{lstlisting}
\end{adjustbox}

\end{document}